\documentclass[]{spie}  %>>> use for US letter paper
\usepackage{amsmath,amsfonts,amssymb}
\usepackage{graphicx,subcaption}
\usepackage[colorlinks=true, allcolors=blue]{hyperref}
\usepackage{multirow}
\usepackage{array}
\usepackage{xcolor}

\title{Optimizing YOLOv7 for Semiconductor Defect Detection}

\author[a,b]{Enrique Dehaerne}
\author[b]{Bappaditya Dey}
\author[b]{Sandip Halder}
\author[b,c]{Stefan De Gendt}
\affil[a]{Dept. of Computer Science, KU Leuven, 3001 Leuven, Belgium}
\affil[b]{Interuniversity Microelectronics Centre (imec), 3001 Leuven, Belgium}
\affil[c]{Dept. of Chemistry, KU Leuven, 3001 Leuven, Belgium}

\authorinfo{Send correspondence to Enrique Dehaerne, e-mail: enrique.dehaerne@imec.be}

\begin{document} 
\maketitle

\begin{abstract}
The field of object detection using Deep Learning (DL) is constantly evolving with many new techniques and models being proposed. YOLOv7 is a state-of-the-art object detector based on the YOLO family of models which have become popular for industrial applications. One such possible application domain can be semiconductor defect inspection.
The performance of any machine learning model depends on its hyperparameters. Furthermore, combining predictions of one or more models in different ways can also affect performance.
In this research, we experiment with YOLOv7, a recently proposed, state-of-the-art object detector, by training and evaluating models with different hyperparameters to  investigate which ones improve performance in terms of detection precision for semiconductor line space pattern defects. The base YOLOv7 model with default hyperparameters and Non Maximum Suppression (NMS) prediction combining outperforms all RetinaNet models from previous work in terms of mean Average Precision (mAP). We find that vertically flipping images randomly during training yields a 3\% improvement in the mean AP of all defect classes. Other hyperparameter values improved AP only for certain classes compared to the default model. Combining models that achieve the best AP for different defect classes was found to be an effective ensembling strategy. Combining predictions from ensembles using Weighted Box Fusion (WBF) prediction gave the best performance. The best ensemble with WBF improved on the mAP of the default model by 10\%.
\end{abstract}


\section{Introduction}\label{sect:introduction}
Automatic defect inspection is important for high-yield and reduced engineering time/cost. Techniques such as scanning electron microscopy (SEM) can produce high-resolution images for inspecting defects at the nanometer scale. The inherently noisy nature of SEM images and continuous transitions to smaller nodes causes traditional, rule-based image processing methods for defect inspection to fail \cite{Dey_2022_retinanet}. This has spurred interest in machine learning, particularly deep learning (DL), methods that are able to handle noise and changes in scale better \cite{imoto2018transferlearning, cheon2019waferdefect, wang2021defectgan, Dey_2022_retinanet, icecs_benchmark, sem_yolov5}.

In this study, we focus on defect detection, which involves localizing and classifying defect instances, in SEM images. Dey et al.\cite{Dey_2022_retinanet} showed that a DL framework based on RetinaNet detects semiconductor defects more robustly than rule-based image processing techniques. DL-based object detection is a very popular research topic with new techniques and models being proposed constantly. YOLOv7\cite{yolov7} is a state-of-the-art DL-based object detection model based on the YOLO\cite{yolo_og} family of models which have become popular for industrial applications. 

This study shows that the base YOLOv7 model architecture with default hyperparameters outperforms multiple RetinaNet models. To improve detection performance, many YOLOv7 models with different hyperparameters are systematically trained and evaluated. Furthermore, two detection prediction combination techniques, Non-Maximum Suppression (NMS) and Weighted Box Fusion (WBF), with different model ensemble strategies are compared. The findings from these experiments provide practical insights for DL-based object detection for semiconductor defect detection. 

\section{Dataset}\label{sect:dataset}
\begin{figure}
    \centering
    \includegraphics[width=\textwidth]{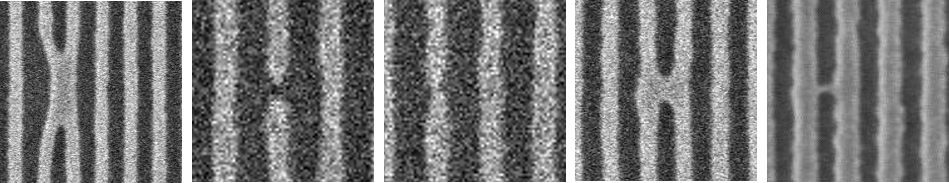}
    \caption{Examples of (from left to right) line collapse, gap, probable gap, bridge, and microbridge line space pattern defects.}
    \label{fig:class_examples}
\end{figure}

A collection of 1324 SEM wafer after-developing-inspection images with 3265 defect instances from Dey et al. \cite{Dey_2022_retinanet} is reused in this study. The images are of line space patterns with line collapse, gap, probable gap (or p-gap), bridge, or microbridge defects. Examples of each of these five defect types are shown in Figure \ref{fig:class_examples}. Table \ref{tab:dataset} shows statistics of the dataset including the number of images and defect instances in the train, validation, and test splits.

\begin{table}[h]
    \caption{SEM image dataset statistics for each split.}
    \label{tab:dataset}
    \begin{center}
    \begin{tabular}{|c|c|c|c|}
        \hline
        \textbf{Sample counts} & \textbf{Train} & \textbf{Validation} & \textbf{Test} \\ 
        \hline
        Line collapse & 550 & 66 & 76 \\ 
        Bridge & 238 & 19 & 17 \\
        Microbridge & 380 & 47 & 78 \\
        Gap & 1046 & 156 & 174 \\ 
        Probable Gap & 315 & 49 & 54 \\ 
        \hline
        Total instances & 2529 & 337 & 399 \\ 
        \hline
        Total images & 1053 & 117 & 154 \\ 
        \hline
    \end{tabular}
    \end{center}
\end{table}

\section{Methodology}\label{sect:methodology}
\begin{figure}
\centering
    \begin{subfigure}{0.32\linewidth}
\includegraphics[width=\linewidth]{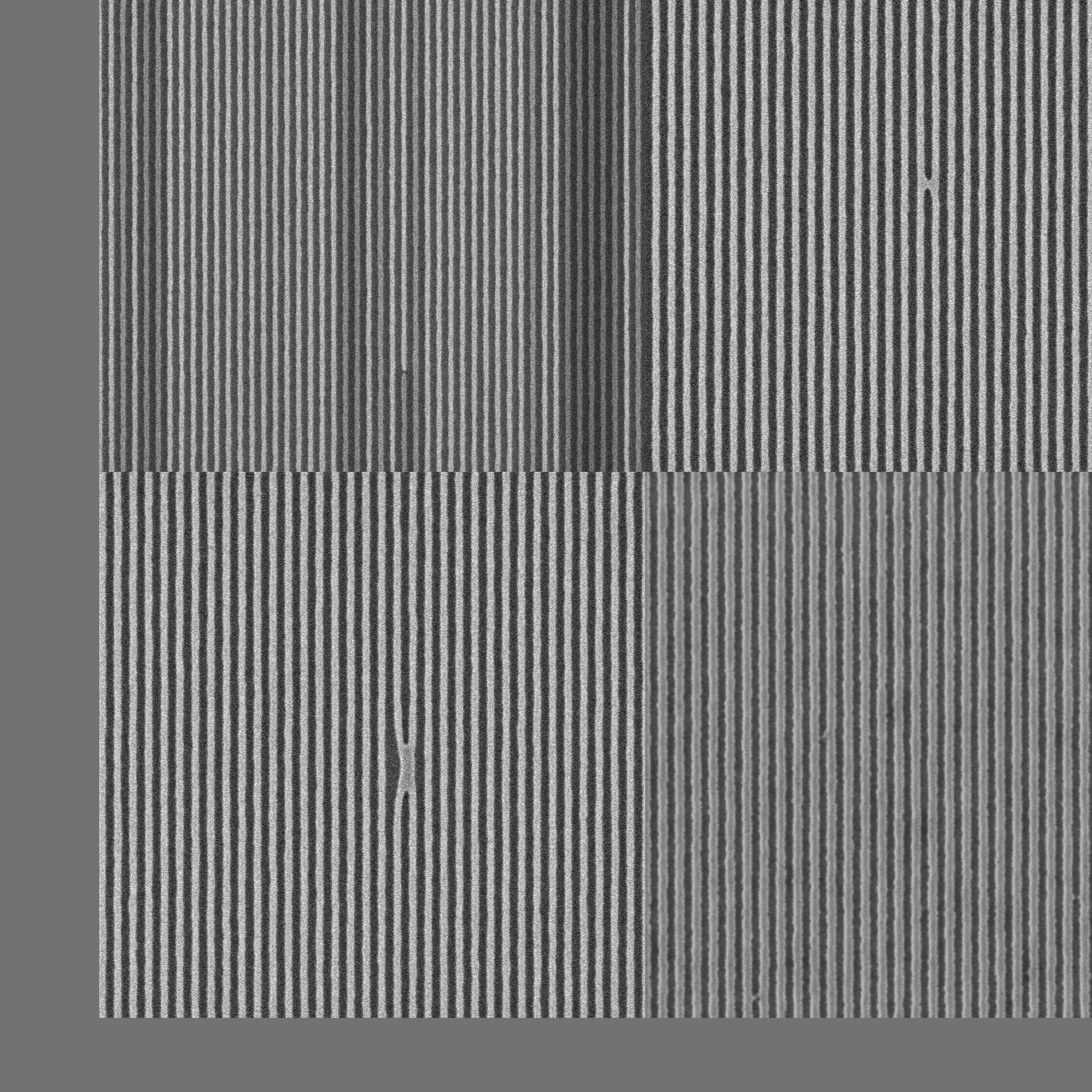} 
    \caption{Mosaic\cite{yolov4_mosaic}}
\label{fig:da_eg_1}
    \end{subfigure}\hfill
    \begin{subfigure}{0.32\linewidth}
\includegraphics[width=\linewidth]{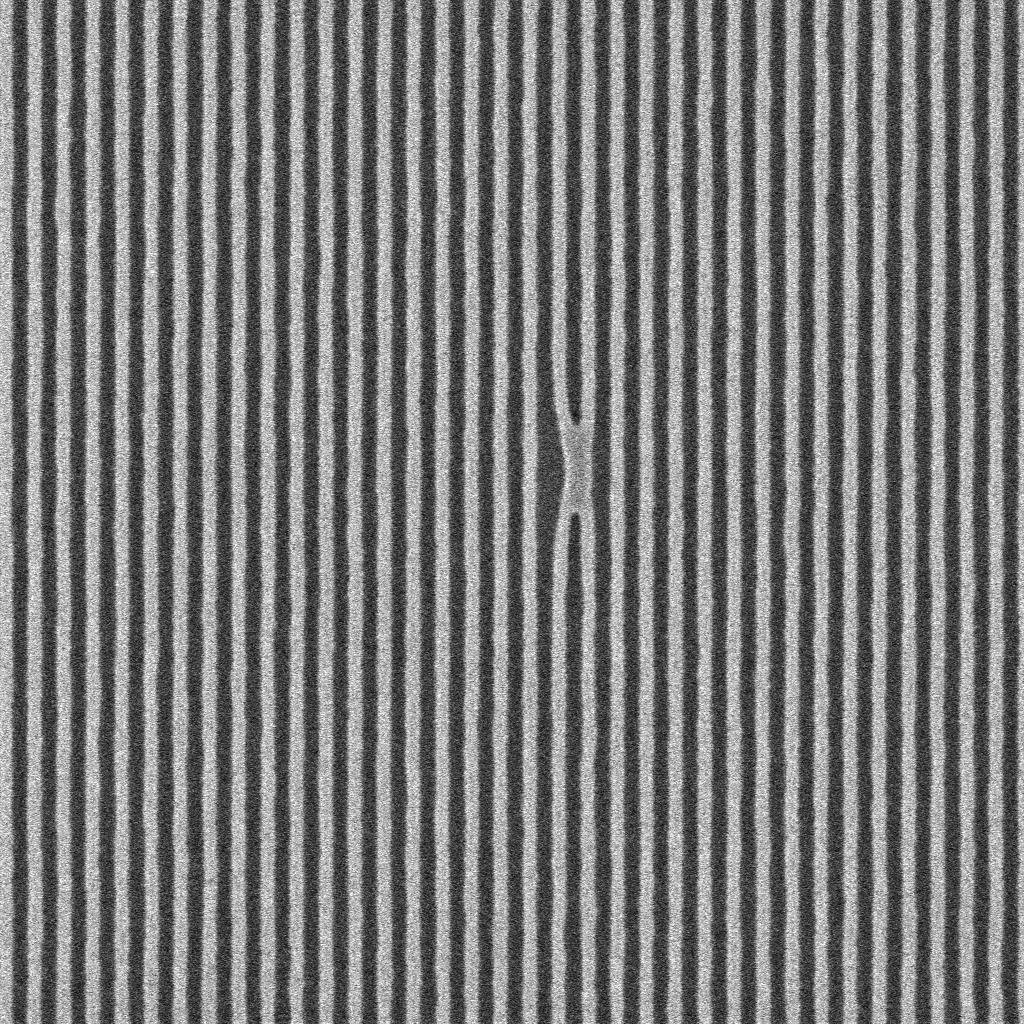}
    \caption{Vertical flip}
\label{fig:da_eg_2}
    \end{subfigure}\hfill
    \begin{subfigure}{0.32\linewidth}
% \includegraphics[width=\linewidth]{example-image-c}
%     \caption{Horizontal flip}
% \label{fig:da_eg_3}
%     \end{subfigure}
%     \begin{subfigure}{0.32\linewidth}
\includegraphics[width=\linewidth]{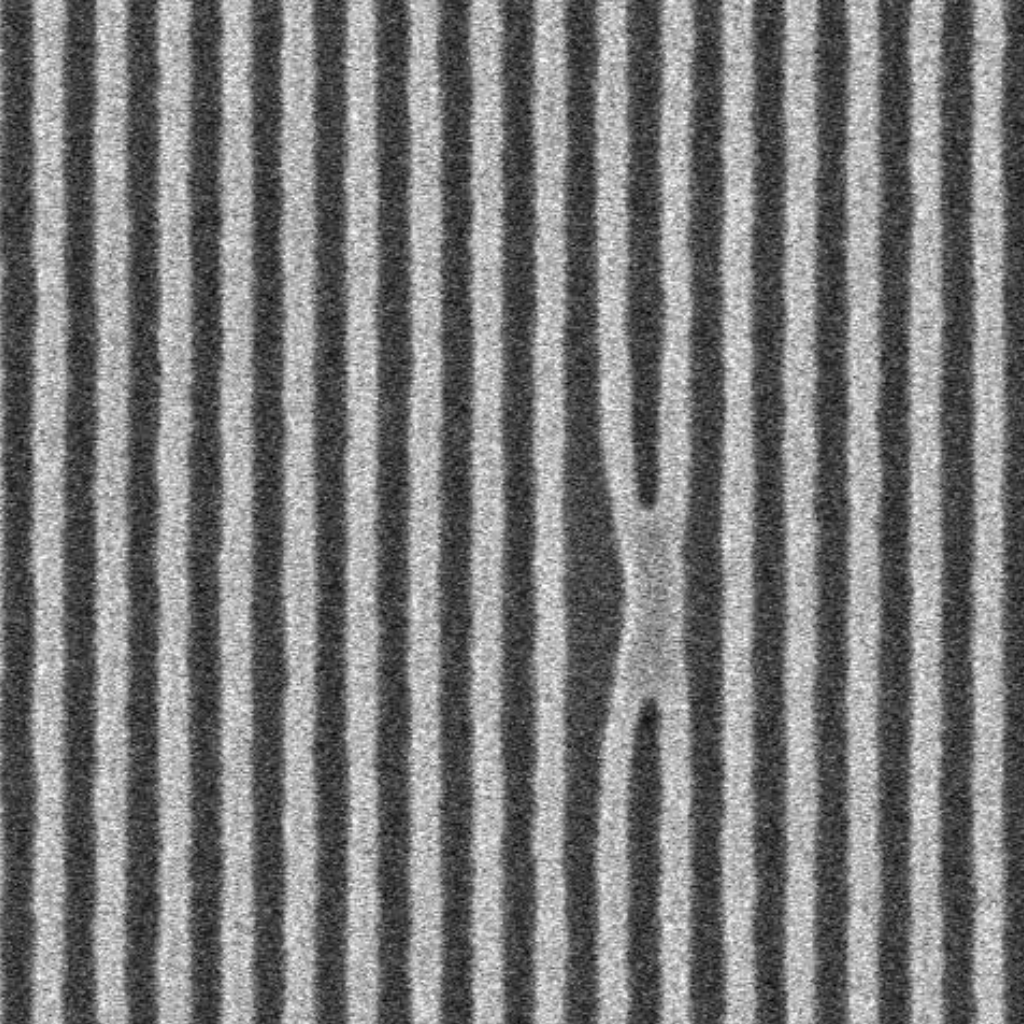} 
    \caption{Scale (+0.15)}
\label{fig:da_eg_4}
    \end{subfigure}\hfill
    \begin{subfigure}{0.32\linewidth}
\includegraphics[width=\linewidth]{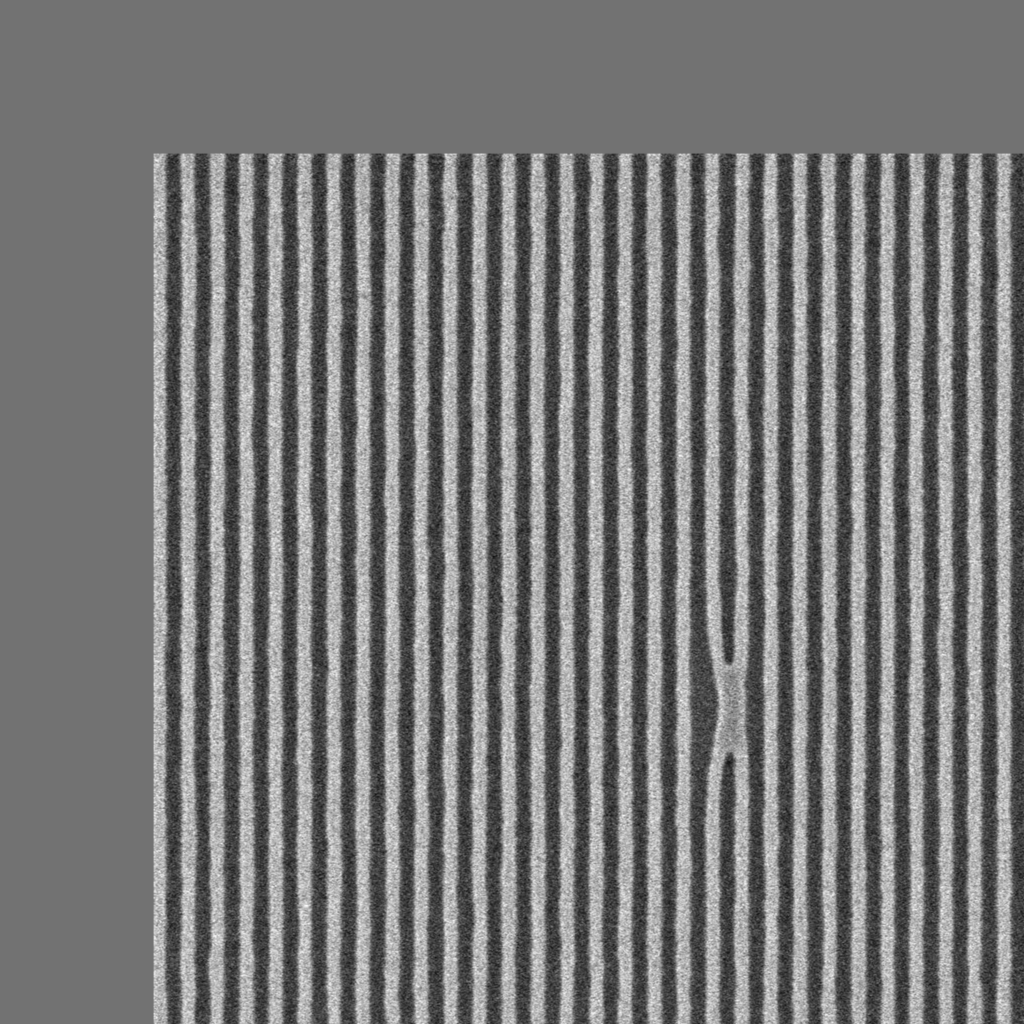}
    \caption{Translation (+0.15)}
\label{fig:da_eg_5}
    \end{subfigure}\hfill
    \begin{subfigure}{0.32\linewidth}
\includegraphics[width=\linewidth]{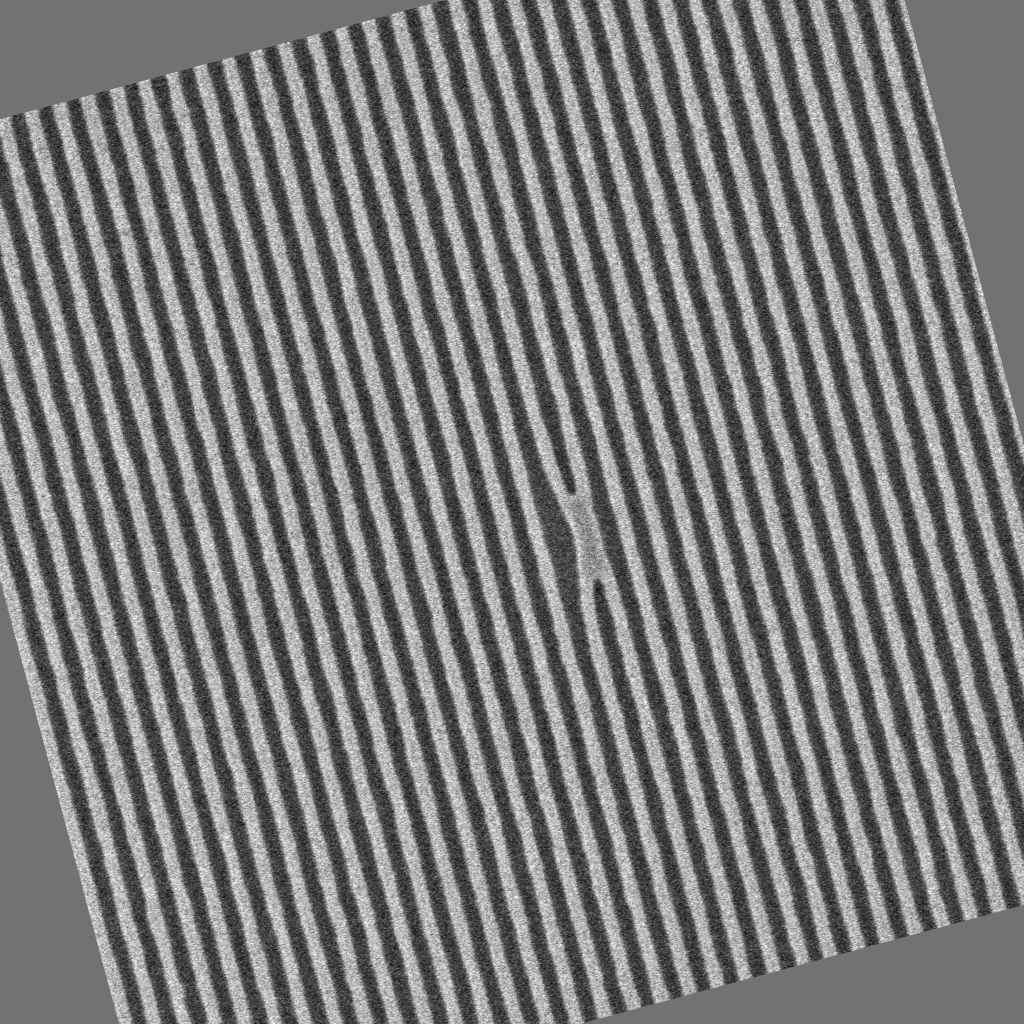}
    \caption{Angle (15 degrees)}
\label{fig:da_eg_6}
    \end{subfigure}
    \begin{subfigure}{0.32\linewidth}
\includegraphics[width=\linewidth]{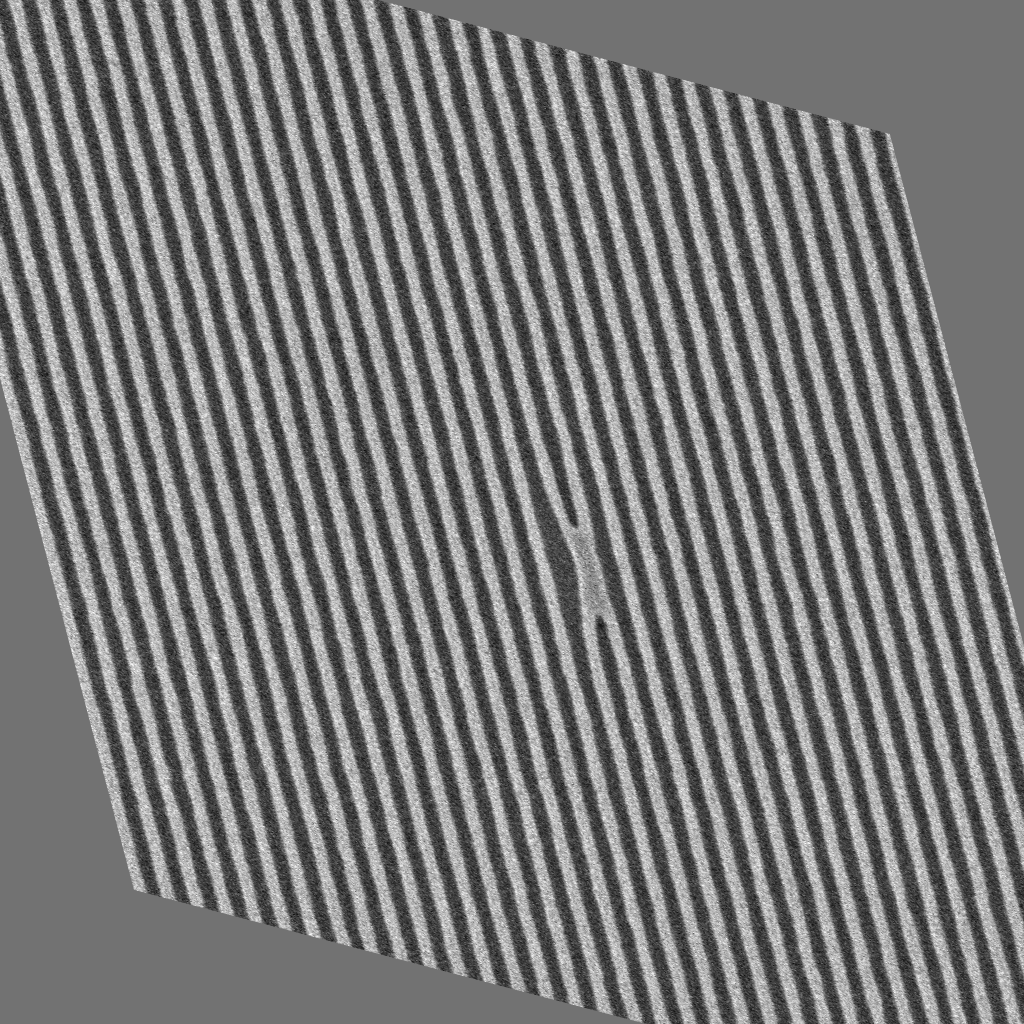}
    \caption{Shear (+15 degrees)}
\label{fig:da_eg_6}
    \end{subfigure}
    \begin{subfigure}{0.32\linewidth}
\includegraphics[width=\linewidth]{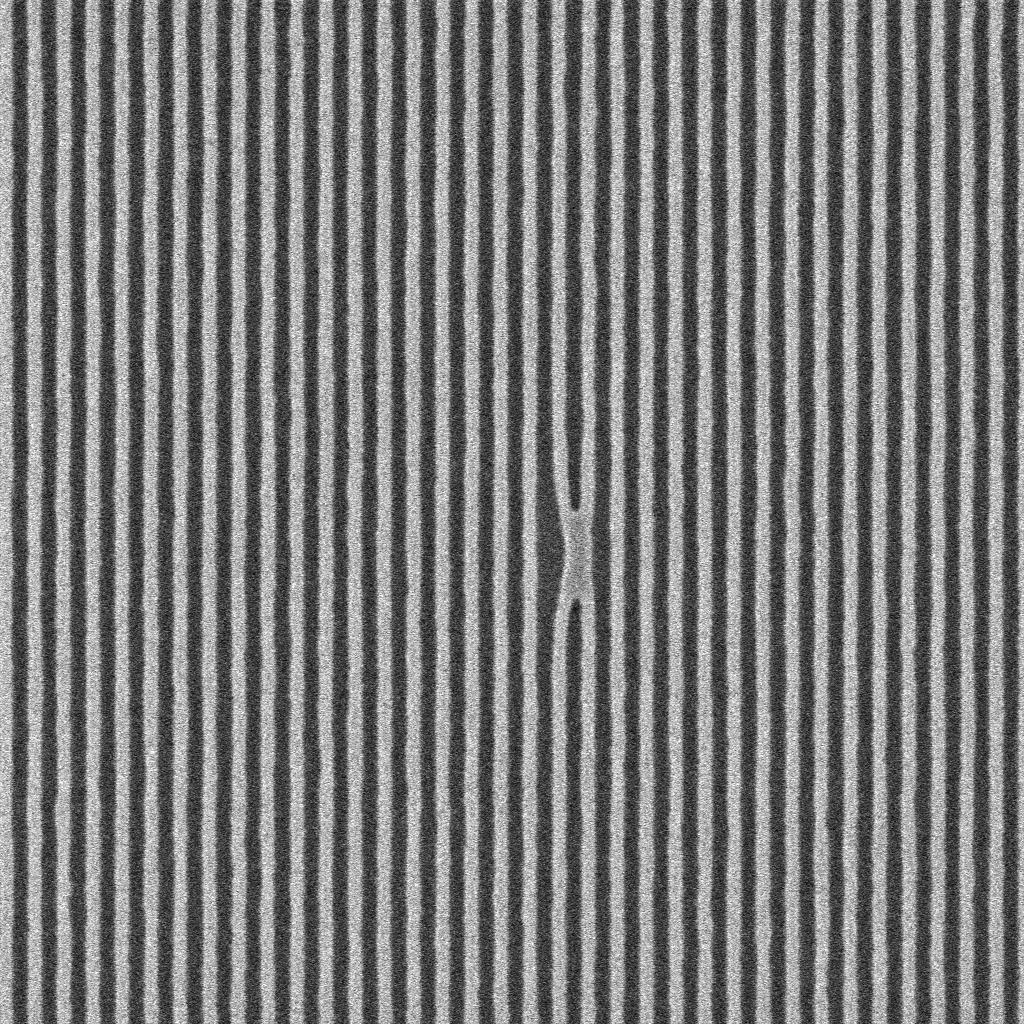} 
    \caption{Hue (+0.5)}
\label{fig:da_eg_7}
    \end{subfigure}\hfill
    \begin{subfigure}{0.32\linewidth}
\includegraphics[width=\linewidth]{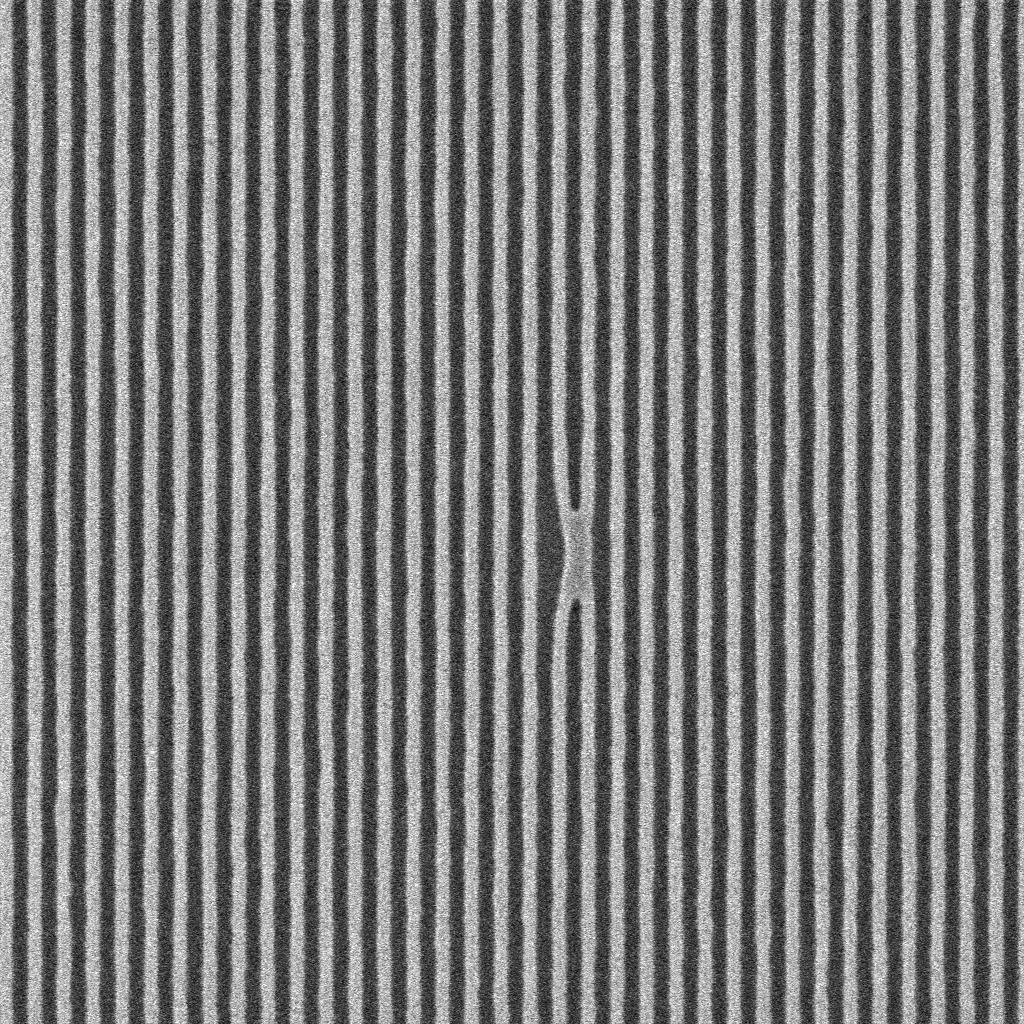}
    \caption{Saturation (+0.5)}
\label{fig:da_eg_8}
    \end{subfigure}\hfill
    \begin{subfigure}{0.32\linewidth}
\includegraphics[width=\linewidth]{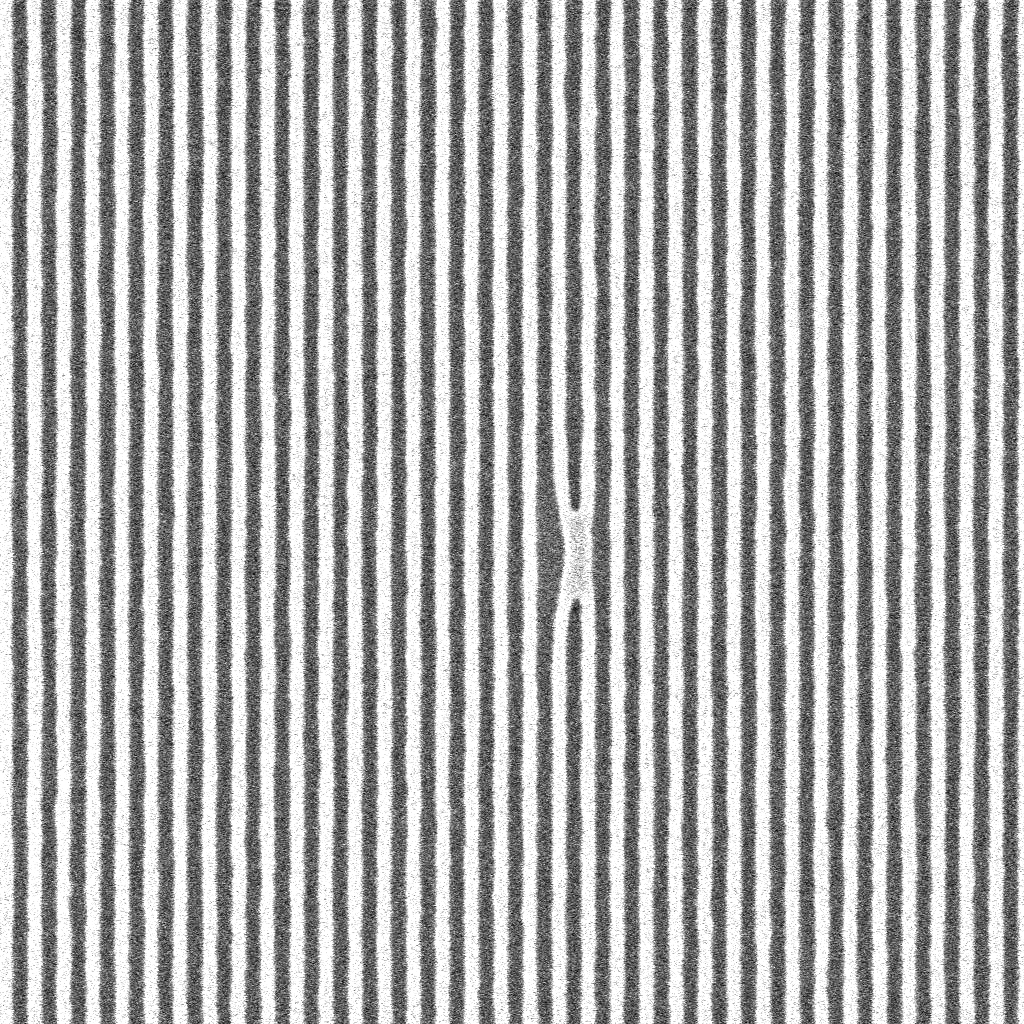}
    \caption{Value (+0.5)}
\label{fig:da_eg_9}
    \end{subfigure}
\caption{Data augmentation examples.}
    \label{fig:da_eg}
\end{figure}

First, the YOLOv7\cite{yolov7} base model was trained using the default hyperparameters. Hyperparameters that were hypothesized to noticeably affect the detection performance of the model were chosen for experimentation. These can be categorized as either ``weights \& learning'' or ``data augmentation'' hyperparameters. The former affects the number of weights of the model or how they are updated. The latter modifies the input images during training. The models were trained so that one of these hyperparameters had been assigned a value different from that of its default value. The hyperparameters and the default and modified values are shown in Table \ref{tab:hyperparameters}. Examples of images after applying each of the chosen data augmentations operations are shown in Figure \ref{fig:da_eg}.
% TODO add more information about both types of hyperparameters

\begin{table}[h]
    \centering
    \caption{Hyperparameters selected for experimentation with their default and chosen modified values.}
    \label{tab:hyperparameters}
    \begin{tabular}{|>{\centering\arraybackslash}m{65pt}|>{\centering\arraybackslash}m{120pt}|c|c|c|}
    \hline
        \textbf{Type} & \textbf{Hyperparameter} & \textbf{Default} & \textbf{Modified (1)} & \textbf{Modified (2)} \\ \hline\hline
        \multirow{9}{50pt}{Weights \& learning}
        & Anchor threshold & 4 & 9 & 13 \\ \cline{2-5}
        & Number of anchors & 3 & 9 & 13\\ \cline{2-5}
        & IOU threshold & 0.2 & 0.5 & 0.75 \\ \cline{2-5}
        & Object loss gain & 0.7 & 0.25 & 0.5 \\ \cline{2-5}
        & Class loss gain & 0.3 & 0.1 & 0.5 \\ \cline{2-5}
        & Box loss gain & 0.05 & 0.1 & 0.25 \\ \cline{2-5}
        & Focal-loss gamma & 0.0 & 1.0 & 1.5 \\ \cline{2-5}
        & Freeze backbone layers & First layer only & First 25 layers & All 50 layers\\ \cline{2-5}
        & Model size & Base & Tiny & Base-X \\ \hline\hline
        \multirow{8}{50pt}[-1em]{Data augmentation}
        & Vertical flipping (probability) & 0.0 & 0.5 & - \\ \cline{2-5}
        & Horizontal flipping (probability) & 0.5 & 0.0 & - \\ \cline{2-5}
        & Mosaic\cite{yolov4_mosaic} & 1.0 & 0.0 & 0.5 \\ \cline{2-5}
        & Scale (+/- gain)  & 0.5 & 0.25 & 0.75 \\ \cline{2-5}
        & Translation (+/- fraction) & 0.2 & 0.0 & 0.5 \\ \cline{2-5}
        & Angle (+/- degrees) & 0 & 45 & 90 \\ \cline{2-5}
        & Shear (+/- degrees) & 0 & 15 & 30 \\ \cline{2-5}
        & HSV (fraction) & 0.015/0.7/0.4  (h/s/v) & 0.0 (all) & 1.0 (all) \\ \hline
        % \multirow{3}{50}[-1em]{Prediction combination} 
        % & Single model \hspace{1cm} (default hyperparams) & NMS & WBF & - \\ \cline{2-5}
        % & Ensemble different sizes (default hyperparameters) & NMS & WBF & - \\ \cline{2-5}
        % & Ensemble different hyperparameters & NMS & WBF & - \\ \hline
    \end{tabular}
\end{table}

\begin{figure}
\centering
    \begin{subfigure}{0.32\linewidth}
\includegraphics[width=\linewidth]{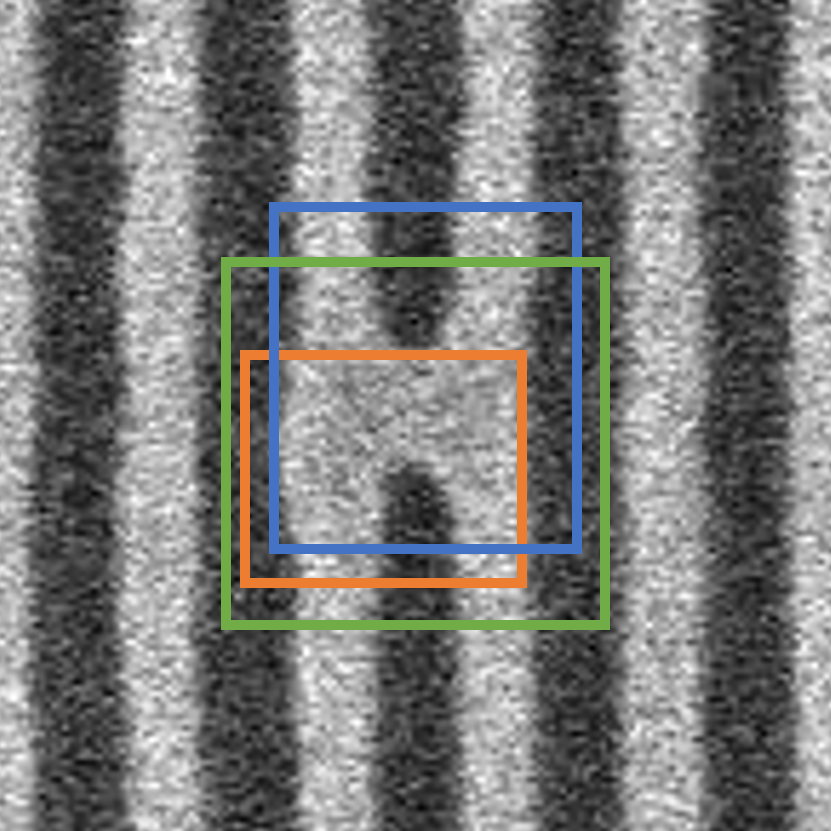} 
    \caption{Predictions before combination}
\label{fig:pred_1}
    \end{subfigure}\hfill
    \begin{subfigure}{0.32\linewidth}
\includegraphics[width=\linewidth]{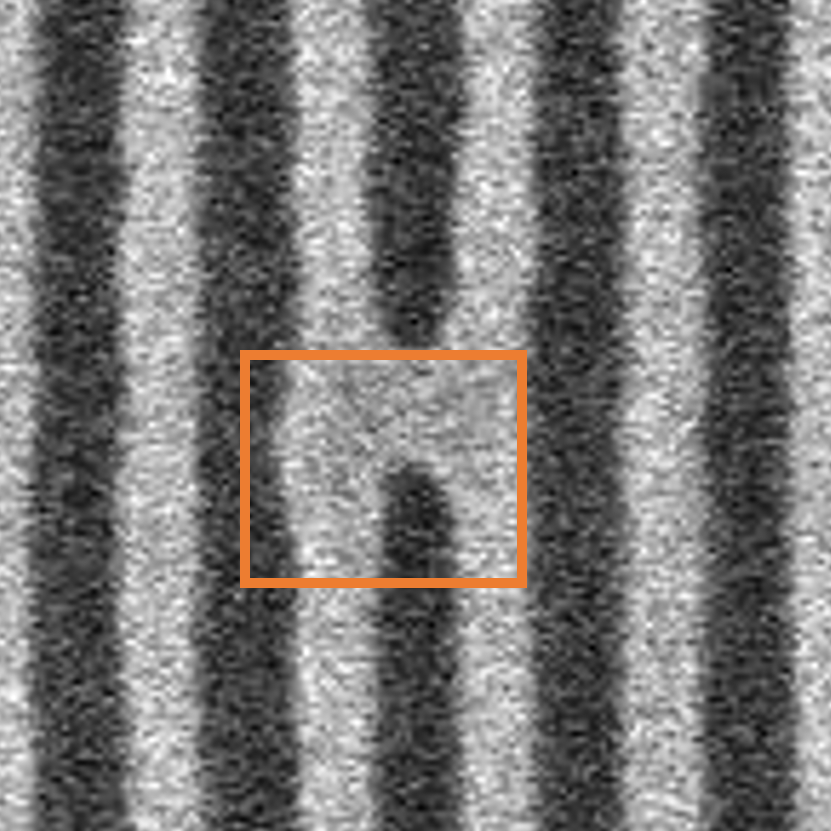}
    \caption{NMS result}
\label{fig:pred_2}
    \end{subfigure}\hfill
    \begin{subfigure}{0.32\linewidth}
\includegraphics[width=\linewidth]{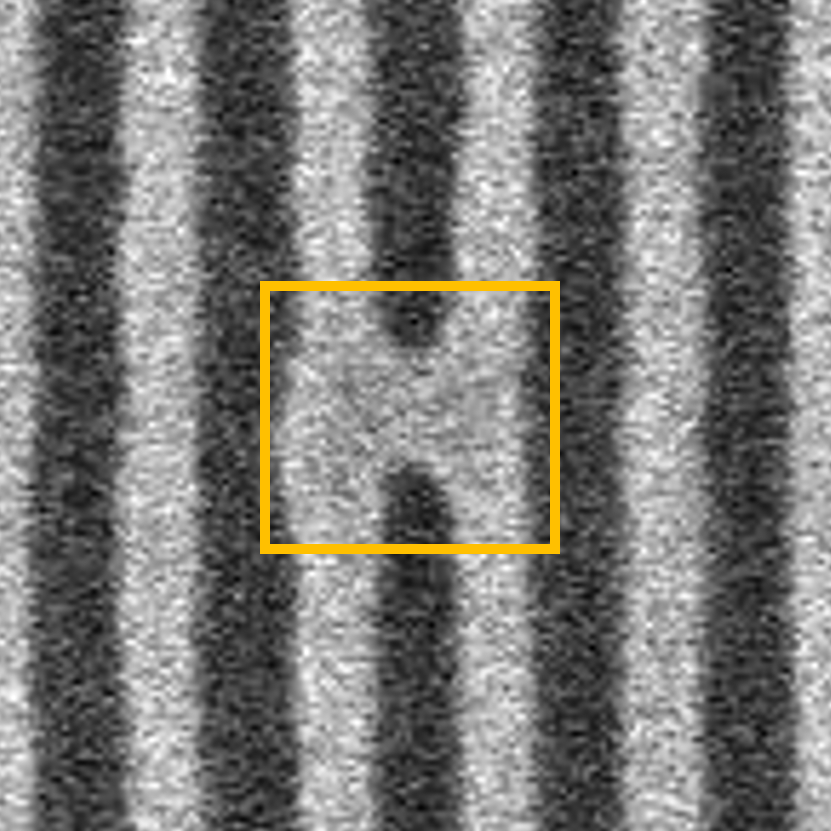}
    \caption{WBF result}
\label{fig:pred_3}
    \end{subfigure}
\caption{Example predictions for a bridge defect (\ref{fig:pred_1}) and corresponding final prediction examples after applying NMS (\ref{fig:pred_2}) and WBF (\ref{fig:pred_3}) combination algorithms. In this example, the orange prediction has the largest confidence score.}
    \label{fig:prediction_combos}
\end{figure}

Detection models like YOLOv7 often predict multiple bounding boxes for the same instance. Therefore, combining these predicted boxes into a single final prediction is considered to be an important step for object detection. The most common prediction combination method of accomplishing is NMS. NMS groups predicted boxes for a class together when their intersection-over-unions are above a given threshold and removes all boxes except for the one with the highest confidence score. A more advanced method is WBF \cite{solovyev2021weighted} which takes a weighted average of each box in a group to create the final prediction box. Figure \ref{fig:prediction_combos}, shows example predictions on a defect and how the NMS and WBF methods would combine the predictions. NMS is used by default for all models trained. We experiment with using WBF for the default model as well as ensemble models.

\begin{figure}
    \centering
    \includegraphics[width=\textwidth]{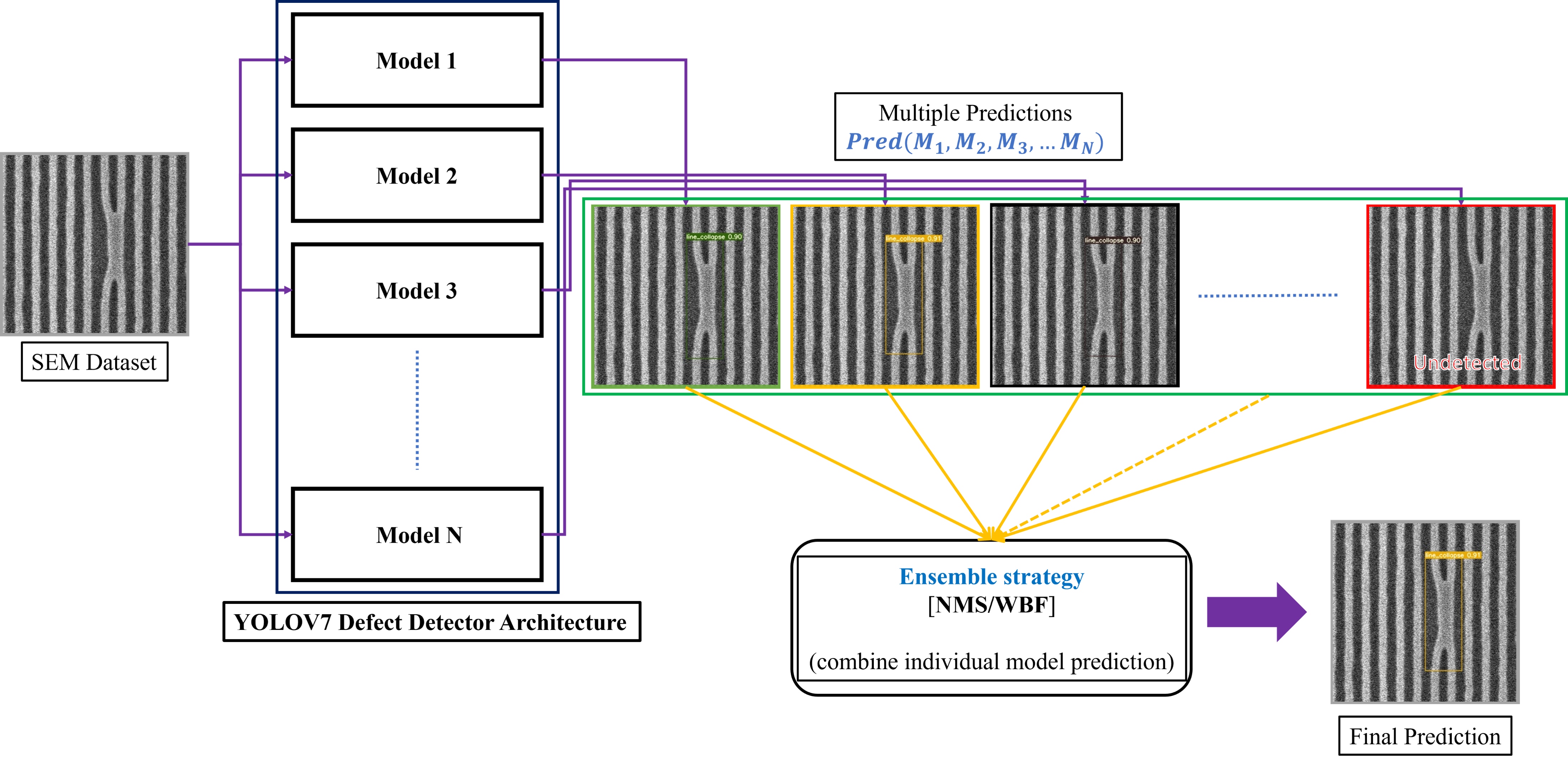}
    \caption{The ensemble framework generalized to N models. The final prediction of the ensemble model reflects the detection with the highest confidence score (NMS).}
    \label{fig:ensemble_framework}
\end{figure}

The best mean Average Precision (mAP) in the study by Dey et al. was achieved by an ensemble model consisting of three RetinaNet models with ResNet backbones of different sizes\cite{Dey_2022_retinanet}. In addition to an ensemble model of YOLOv7 models of different sizes (Tiny, Base, and Base-X), we also ensembled models trained with different hyperparameters that achieve the best AP for different defect classes. Figure \ref{fig:ensemble_framework} depicts how ensemble models work on a conceptual level.

% Figure showing nms and wbf in action on an image

\section{Experimental Setup}\label{sect:experiment_setup}
Code from the official YOLOv7\footnote{\url{https://github.com/WongKinYiu/yolov7}} and WBF\footnote{\url{https://github.com/ZFTurbo/Weighted-Boxes-Fusion}} GitHub repositories were used to implement the experiments. All models were trained with a batch size of 2 for 200 epochs. At each epoch, evaluation was performed on the validation split. For each model, the trained checkpoint that achieved the best mAP at an IOU threshold of 0.5 was chosen for evaluation on the test split. All experiments were run on a Lambda Vector workstation with an NVIDIA GeForce RTX 3070 GPU.

\section{Results \& Discussion}\label{sect:results}
The base YOLOv7 model with default hyperparameters achieves an mAP score of 0.790. This outperforms the best RetinaNet\cite{retinanet} models from Dey et al.\cite{Dey_2022_retinanet} (0.787/0.775/0.788 for ResNet50/101/152 backbones\cite{resnet}, respectively). 
% This supports the idea that model architecture improvements in popular object detection literature, which is measured on natural image object detection benchmark datasets such as COCO\cite{}, will also translate to improvements in semiconductor defect detection in SEM images. 
As will be shown in the rest of this section, the default YOLOv7, with hyperparameters optimized for natural image object detection, achieves a better mAP score than all but one of the models trained with different hyperparameters.  

Table \ref{tab:wnl test results} shows the per-class AP and overall mAP on the test images for models with different ``weights \& learning'' hyperparameter values. All models trained with a modified hyperparameter achieve lower mAPs than the default model. This shows that manually tuning hyperparameters for a certain dataset is difficult. Future work should instead employ hyperparameter optimization algorithms such as genetic algorithms, grid search, or bayesian optimization\cite{hpo_survey,sem_yolov5}. Certain models do achieve better APs for the bridge and p-gap classes. Most notably, the Tiny model achieves a 36\% better bridge AP than the default model.

\begin{table}[h]
    \centering
    \caption{Per-class AP and overall mAP on the test images for all YOLOv7 base models with different ``weights \& learning'' hyperparameter values. Bold values indicate that the default model's value is the best or if the value is better than the default model's results.}
    \label{tab:wnl test results}
    \begin{tabular}{|c|c||c|c|c|c|c|c|}
        \hline
        \multirow{2}{*}{\textbf{Hyperparameter}} & \multirow{2}{*}{\textbf{Value}} & \multicolumn{6}{c|}{\textbf{AP (@0.5 IOU)}} \\
        \cline{3-8}
        & & \textbf{microbridge} & \textbf{gap} & \textbf{bridge} & \textbf{line collapse} & \textbf{p-gap} & \textbf{All (mAP)} \\
        \hline\hline
        \multicolumn{2}{|c||}{Default} & \textbf{0.873} & \textbf{0.967} & 0.602 & \textbf{1.000} & 0.508 & \textbf{0.790} \\
        \hline
        \multirow{2}*{Anchor threshold} & 9 & 0.806 & 0.950 & \textbf{0.639} & 1.000 & \textbf{0.529} & 0.785 \\
        \cline{2-8}
        & 13 & 0.792 & 0.958 & 0.537 & 1.000 & 0.238 & 0.705 \\
        \hline

        \multirow{2}*{Anchors} & 9 & 0.726 & 0.948 & 0.587 & 1.000 & 0.167 & 0.686 \\
        \cline{2-8}
        & 13 & 0.766 & 0.948 & 0.477 & 0.000 & 0.103 & 0.574 \\ % line collapse given by script was -1.000
        \hline
        
        \multirow{2}*{IOU threshold} 
        & 0.1 & 0.737 & 0.950 & 0.590 & 1.000 & 0.150 & 0.685 \\
        \cline{2-8}
        & 0.75 & 0.807 & 0.959 & 0.609 & 1.000 & 0.163 & 0.708 \\
        \hline

        \multirow{2}*{Object loss gain} 
        & 0.25 & 0.754 & 0.949 & 0.581 & 1.000 & 0.274 & 0.712 \\
        \cline{2-8}
        & 0.5 & 0.800 & 0.959 & \textbf{0.750} & 1.000 & 0.275 & 0.757 \\
        \hline

        \multirow{2}*{Class loss gain} 
        & 0.1 & 0.737 & 0.950 & 0.590 & 1.000 & 0.150 & 0.685 \\ \cline{2-8}
        & 0.5 & 0.803 & 0.958 & 0.583 & 1.000 & 0.457 & 0.760 \\
        \hline

        \multirow{2}*{Box loss gain} 
        & 0.1 & 0.762 & 0.959 & 0.562 & 1.000 & 0.106 & 0.678 \\
        \cline{2-8}
        & 0.5 & 0.800 & 0.959 & 0.750 & 1.000 & 0.275 & 0.757 \\
        \hline

        \multirow{2}*{Focal-loss gamma} 
        & 1.0 & 0.635 & 0.890 & 0.652 & 0.980 & 0.000 & 0.631 \\
        \cline{2-8}
        & 1.5 & 0.581 & 0.851 & 0.505 & 1.000 & 0.000 & 0.587 \\
        \hline

        \multirow{2}*{Freeze layers} 
        & 25 & 0.712 & 0.919 & 0.584 & 1.000 & 0.247 & 0.693 \\
        \cline{2-8}
        & 50 & 0.745 & 0.949 & 0.579 & 1.000 & 0.139 & 0.682 \\
        \hline

        \multirow{2}*{Model size} 
        & Tiny & 0.746 & 0.960 & \textbf{0.819} & 1.000 & 0.281 & 0.761 \\
        \cline{2-8}
        & Base-X & 0.821 & 0.960 & 0.515 & 1.000 & 0.191 & 0.697 \\
        \hline
        
    \end{tabular}
\end{table}

\begin{table}[h]
    \centering
    \caption{Per-class AP and overall mean AP on the test images for all YOLOv7 base models with different ``data augmentation'' hyperparameter values. Bold values indicate that the default model's value is the best or if the value is better than the default model's results.}
    \label{tab:da test results}
    \begin{tabular}{|c|c||c|c|c|c|c|c|}
        \hline
        \multirow{2}{*}{\textbf{Hyperparameter}} & \multirow{2}{*}{\textbf{Value}} & \multicolumn{6}{c|}{\textbf{AP (@0.5 IOU)}} \\
        \cline{3-8}
        & & \textbf{microbridge} & \textbf{gap} & \textbf{bridge} & \textbf{line collapse} & \textbf{p-gap} & \textbf{All (mAP)} \\
        \hline\hline
        \multicolumn{2}{|c||}{Default} & \textbf{0.873} & 0.967 & 0.602 & \textbf{1.000} & 0.508 & 0.790 \\
        \hline
        Vertical Flipping & 0.5 & 0.709 & 0.960 & \textbf{0.790} & 1.000 & \textbf{0.604} & \textbf{0.812} \\ \hline
        Horizontal Flipping & 0.0 & 0.722 & 0.959 & \textbf{0.718} & 1.000 & 0.507 & 0.781\\ \hline

        \multirow{2}*{Mosaic\cite{yolov4_mosaic}}
        & 0.0 & 0.647 & 0.952 & 0.581 & 1.000 & 0.030 & 0.642 \\ \cline{2-8}
        & 0.5 & 0.780 & 0.949 & 0.589 & 1.000 & 0.277 & 0.719 \\ 
        \hline
        
        \multirow{2}*{Scale}
        & 0.25 & 0.822 & 0.949 & 0.437 & 1.000 & 0.288 & 0.699 \\
        \cline{2-8}
        & 0.75 & 0.758 & 0.939 & 0.634 & 1.000 & 0.133 & 0.693 \\
        \hline

        \multirow{2}*{Translation} 
        & 0.0 & 0.784 & \textbf{0.968} & 0.540 & 1.000 & 0.107 & 0.680 \\
        \cline{2-8}
        & 0.5 & 0.808 & 0.940 & 0.457 & 1.000 & 0.195 & 0.680 \\
        \hline

        \multirow{2}*{Angle} 
        & 45 & 0.633 & 0.959 & \textbf{0.912} & 1.000 & 0.268 & 0.754 \\
        \cline{2-8}
        & 90 & 0.597 & 0.899 & 0.745 & 1.000 & 0.055 & 0.659 \\
        \hline

        \multirow{2}*{Shear} 
        & 15 & 0.779 & 0.967 & 0.548 & 1.000 & 0.277 & 0.714 \\
        \cline{2-8}
        & 30 & 0.785 & \textbf{0.968} & 0.575 & 1.000 & 0.346 & 0.735 \\
        \hline

        \multirow{2}*{HSV} 
        & 0.0 & 0.781 & 0.949 & 0.586 & 1.000 & 0.326 & 0.729 \\
        \cline{2-8}
        & 1.0 & 0.677 & 0.949 & 0.584 & 1.000 & 0.197 & 0.681 \\
        \hline
        
    \end{tabular}
\end{table}

Table \ref{tab:da test results} shows the AP results for models with different ``data augmentation'' hyperparameter values. We find that vertically flipping images randomly during training yields a 19\% improvement in the AP for the most difficult defect class, probable gap, and a 3\% improvement in the mean AP of all defect classes. This makes sense since a semiconductor SEM image flipped vertically shares the same characteristics as an image the model might encounter at inference time. Other hyperparameters don't achieve a better mAP score than the default model. However, just like the `` weights \& learning'' hyperparameters, some other data augmentations hyperparameters do improve on AP for certain classes. Notably, tilting images randomly at an angle no greater than 45 degrees achieves the best bridge AP, 0.912, of all models tested.  

Table \ref{tab:ens test results} shows the AP results for different ensembles of models with either NMS or WBF prediction combination methods. The most impressive improvements come from ensembling models and applying WBF. All ensemble models achieve better per-class AP results for all classes and better mAP results than the default model alone. Applying WBF to the default model and to ensembles also improved mAP. Combining predictions from different models of different sizes using NMS and WBF improves mAP by 4\% and 7\%, respectively. The default, vertical flipping, and 45-degree angle models were chosen for the ensemble of models of different hyperparameter values that achieve best per-class performances. These models achieved the best microbridge, probable gap, and bridge results, respectively.
% For the ensembles that combine models trained with different hyperparameters based on best per-class performance, the default, vertical flipping, and 45-degree angle models were chosen since they gave the best microbridge, p-gap, and bridge results, respectively. 
This ensemble achieved the best mAP. Combining predictions from these models combined using WBF improved the mAP of the default model by 10\% and the mAP of the size ensemble using WBF by 3\%. The tradeoff here is reduced per-class APs for stand-out individual class results like for bridge defects where the AP of the model with 45-degree angle data augmentation is 7\% better than the ensemble's bridge AP.
% To investigate whether ensembling based on model size or best per-class performance results in better mAP. The results indicated that ensembling models based on best per-class performance gives better mAP than ensembling based on model size. This 
\begin{table}[h]
    \centering
    \caption{Per-class AP and overall mean AP on the test images for all YOLOv7 ensembles with either NMS or WBF prediction combination methods. Bold values indicate that the default model's value is the best or if the value is better than the default model's results. Blue values indicate the best value.}
    \label{tab:ens test results}
    \begin{tabular}{|>{\centering\arraybackslash}m{85pt}|c||c|c|c|c|c|c|}
        \hline
        \multirow{2}{*}{\textbf{Models}} & \textbf{Prediction} & \multicolumn{6}{c|}{\textbf{AP (@0.5 IOU)}} \\
        \cline{3-8}
        & \textbf{combination} & \textbf{microbridge} & \textbf{gap} & \textbf{bridge} & \textbf{line collapse} & \textbf{p-gap} & \textbf{mAP} \\
        \hline\hline
        \multirow{2}*{Default} & NMS &  0.873 & 0.967 & 0.602 & \textbf{1.000} & 0.508 & 0.790 \\
        \cline{2-8}
        & WBF & 0.709 & 0.960 & \textbf{0.790} & 1.000 & \textbf{0.604} & \textbf{0.812} \\ \hline
        % \multirow{2}*{Ensemble different sizes}
        \multirow{2}={Default, Tiny, Base-X}
        & NMS & 0.849 & \textbf{0.968} & \textbf{0.760} & 1.000 & \textbf{0.546} & \textbf{0.825} \\ \cline{2-8}
        & WBF & 0.852 & \textbf{0.968} & \textbf{0.823} & 1.000 & \textbf{0.565} & \textbf{0.842} \\ 
        \hline
        % def, vertical flip, angle
        \multirow{2}={Default, Vertical Flipping, Angle}
        % Ensemble different hyperparams (w Angle)
        & NMS & \textbf{0.877} & \textcolor{blue}{\textbf{0.969}} & \textbf{0.809} & 1.000 & \textbf{0.634} & \textbf{0.858} \\
        \cline{2-8}
         & WBF & \textcolor{blue}{\textbf{0.878}} & \textcolor{blue}{\textbf{0.969}} & \textcolor{blue}{\textbf{0.850}} & 1.000 & \textcolor{blue}{\textbf{0.642}} & \textcolor{blue}{\textbf{0.868}} \\
        \hline
        % def, vertical flip, tiny
        % \multirow{2}={Default, Vertical Flipping, Tiny}
        % & NMS & 0.848 & \textbf{0.968} & \textbf{0.785} & 1.000 & \textbf{0.628} & \textbf{0.846} \\
        % \cline{2-8}
        % & WBF & 0.853 & \textcolor{blue}{\textbf{0.969}} & \textbf{0.834} & 1.000 & \textcolor{blue}{\textbf{0.643}} & \textbf{0.860} \\
        % \hline
    \end{tabular}
\end{table}

% Maybe just work this section in the results and discussion section?
% \section{Limitations \& Future Work}\label{sect:limitations_future_work}
% Only changing one hyperparameter at a time, changing multiple different combinations is a good idea but is computationally expensive. Future work should apply hyperparameter optimization algorithms. 

\section{Conclusion}\label{sect:conclusion}
In this paper, we optimized the YOLOv7 model architecture for semiconductor defect detection by manually modifying hyperparameters and combining predictions from different models. Flipping images vertically during training was the only hyperparameter shown to improve mean Average Precision (mAP) compared to a model trained with default hyperparameters. However, modifying some other hyperparameters resulted in better AP scores for particular defect classes. Combining predictions from models trained with different hyperparameter values that achieved the best APs for different classes significantly improved mAP. Using
the Weighted Box Fusion (WBF) method for combining predictions was shown to improve the results. Ultimately, the best per-class AP ensembling strategy with WBF achieves an mAP that is 10\% better than using a single YOLOv7 model with default hyperparameters. Future work could improve on the results found in this study by using advanced hyperparameter optimization techniques.

% References
\bibliography{main} % bibliography data in report.bib
\bibliographystyle{spiebib} % makes bibtex use spiebib.bst

\end{document}